\documentclass{article}
\usepackage{ijcai24}

\usepackage{times}
\usepackage{soul}
\usepackage{url}
\usepackage[hidelinks]{hyperref}
\usepackage[utf8]{inputenc}
\usepackage[small]{caption}
\usepackage{graphicx}
\usepackage{amsmath}
\usepackage{amsthm}
\usepackage{booktabs}
\usepackage{algorithm}
\usepackage{algorithmic}
\usepackage[switch]{lineno}
\usepackage{multirow}
\usepackage{multicol}
\usepackage{amssymb}
\usepackage{bbding}
\usepackage{wasysym}
\usepackage{hyperref}
\title{SEMv3: A Fast and Robust Approach to Table Separation Line Detection}

\author{
Chunxia Qin$^{1}$ \and
Zhenrong Zhang$^{1}$\and
Pengfei Hu$^{1}$\and
Chenyu Liu$^2$\and \\
Jiefeng Ma$^{1}$\And
Jun Du$^1$\footnote{Corresponding author}
\affiliations
$^1$University of Science and Technology of China\\
$^2$iFLYTEK Research\\
\emails
\{cxqin, zzr666, pengfeihu, jfma\}@mail.ustc.edu.cn, \\cyliu7@iflytek.com,
jundu@ustc.edu.cn
}
\begin{document}
\maketitle

\begin{abstract}
Table structure recognition (TSR) aims to parse the inherent structure of a table from its input image. The ``split-and-merge" paradigm is a pivotal approach to parse table structure, where the table separation line detection is crucial. However, challenges such as wireless and deformed tables make it demanding.  In this paper, we adhere to the ``split-and-merge" paradigm and propose SEMv3 (SEM: \textbf{S}plit, \textbf{E}mbed and \textbf{M}erge), a method that is both fast and robust for detecting table separation lines. During the split stage, we introduce a \textbf{K}eypoint \textbf{O}ffset \textbf{R}egression (KOR) module, which effectively detects table separation lines by directly regressing the offset of each line relative to its keypoint proposals. Moreover, in the merge stage, we define a series of merge actions to efficiently describe the table structure based on table grids. Extensive ablation studies demonstrate that our proposed KOR module can detect table separation lines quickly and accurately. Furthermore, on public datasets (e.g. WTW, ICDAR-2019 cTDaR Historical and iFLYTAB), SEMv3 achieves state-of-the-art (SOTA) performance. The code is available at https://github.com/Chunchunwumu/SEMv3.

\end{abstract}

\section{Introduction}
Tables serve as efficient tools for organizing and presenting crucial information in documents. The structure of a table and the content within its cells collaboratively convey the information in the table. Table structure typically refers to the relationships between cells in a table, involving the spanning of rows and columns. Table structure recognition (TSR) aims to extract inherent structure of a table from unstructured tabular data (table image or scanned document). Table structure recognition plays a crucial role in document digitization and intelligent document analysis~\cite{chi2019scitsr,Zheng2021GTE}.

%Tables are widely used in various application scenarios, including science~\cite{chi2019scitsr,zhong2020pubtabnet}, education~\cite{zhang2023semv2}, and finance~\cite{Zheng2021GTE}.

The TSR methods based on deep learning can be broadly categorized into three paradigms: image-to-markup based, region based and split-and-merge based. 
Image-to-markup based TSR methods~\cite{zhong2020pubtabnet,Smock2022pubtables1m} directly generate table structure tokens from table images, which exhibit weaknesses in handling large tables due to the length of generation. 
Region based TSR methods~\cite{long2021wtw,liu2022NCGM} initially detect the cell boxes and then predict the logical adjacency of the detected table cells. However, in detecting table cells, they face challenge in robustly handling cases where visual clues are not obvious, such as wireless tables or blank cells. Split-and-merge based TSR methods~\cite{zhang2023semv2,wang2023tsrformer2,lyu2023gridformer} initially employs a split sub-network to detect table row and column separation lines. Subsequently, these separation lines are intersected to delineate the table grids. A merge sub-network is then utilized to merge the over-split table grids into table cells. Recently, split-and-merge based approach for TSR has attracted a lot of research attention.

\begin{figure}
    \centering
    \includegraphics[width=1\linewidth]{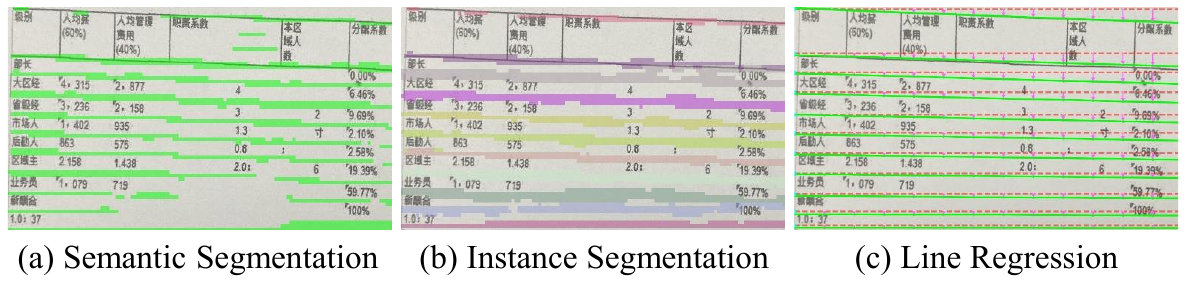}
    \caption{ The three types of table separation detection methods. }
    \label{fig:intro}
\end{figure}
 In the pipeline of split-and-merge based solution, the results of table separation line detection highly impact the accuracy of table structure recognition. Due to the diversity in table styles, accurately detecting table separation lines is challenging. As illustrated in Fig.~\ref{fig:intro}, the prevailing approaches for table separation line detection can be categorized according to the following types. The first type methods~\cite{Schreiber2017DeepDeSRT,zhang2022sem,ma2023RobusTabNet} address the separation line detection through semantic segmentation, which require complex mask-to-line post-processing to parse table separation lines, as shown in Fig.~\ref{fig:intro}(a). The second paradigm conceives the table separation line detection task within the framework of instance segmentation~\cite{zhang2023semv2}, as shown in Fig.~\ref{fig:intro}(b). Each separation line instance has a similar slender appearance. In order to distinguish different instances, position-sensitive instance representation features are important. However, position-sensitive instance information is difficult to learn during the training process based on instance segmentation loss. This leads to less robust separation line detection results compared to our method.
The third paradigm formulates separation line detection as a line regression problem~\cite{lin2022tsrformer}.

% --------------------------------------------------------------------------

% our method
In order to strengthen the robustness of separation line detection, we introduce a new Keypoints Offset Regression module (KOR) in this paper. This module approaches separation line detection from a line regression perspective, as shown in Fig.~\ref{fig:intro}(c). Specifically, KOR represents a separation line as keypoints and regresses the offsets between the line and its keypoint proposals to accurately locate the line.
Compared to other segmentation-based separation line detection methods, our approach eliminates the need for complex mask-to-line post-processing and is more robust. In contrast to previous line regression methods~\cite{lin2022tsrformer,wang2023tsrformer2,lyu2023gridformer}, we do not directly predict the absolute coordinates of line keypoints within the table image. Instead, we predict the offsets from the proposals, effectively incorporating the location prior of the keypoints, which reduces the difficulty of regression.

Following the split stage, the over-split table grids are merged in the merge stage. In order to achieve a precise and effective merging of these table grids, we introduce the concept of a ``merge action" to denote the merging operation for each table grid. We utilize only simple convolutional layers in the construction of the merge module to predict the merging actions of the grids.

% contribution
We amalgamate the Keypoints Offset Regression module (KOR) with the merge module, constituting an end-to-end table structure recognition system denominated SEMv3. Our primary contributions are delineated as follows:
\begin{itemize}
\item We propose Keypoints Offset Regression module (KOR) to robustly and quickly detect the table separation line. KOR represents separation lines with a series of keypoints, locating these keypoints by regressing the offset from proposals to keypoints.
\item We define the merging action of each grid to describe table structure, which can effectively improve the accuracy of merging.
\item Our proposed SEMv3 attains state-of-the-art results on prominent public benchmarks, including ICDAR-2019 cTDaR Historical~\cite{Gao2019icdar2019}, WTW~\cite{long2021wtw}, and iFLYTAB~\cite{zhang2023semv2}. SEMv3 excels in diverse and challenging TSR scenarios, encompassing wireless tables, and deformed tables.
\end{itemize}

\section{Related Work}
\subsection{Image-to-markup Based TSR }
Image-to-markup based TSR methods~\cite{zhong2020pubtabnet,nassar2022tableFormer,Huang2023VAST} use encoder-decoder architecture to generate table structure markups from a table image. However, auto-regressive decoding the structure markups causes error accumulation and inefficient inference. DRCC~\cite{shen2023drcc} adopts a cascaded two-step decoder architecture, predicting row tags first and then cell tags for each row in a semi-autoregressive manner. This approach alleviates the error accumulation problem specific to auto-regressive models. OTSL~\cite{lysak2023optimized} introduces a new table structure representation language, reducing the difficulty of decoding structure tokens. Nevertheless, Image-to-markup based TSR methods require a substantial amount of training data to achieve optimal performance.

\subsection{Region Based TSR}
Region based TSR methods~\cite{xue2019res2tim,zou2020icsp,prasad2020cascadetabnet,xiao2022table} initially identify primitive regions, such as text segments or cell boxes, and subsequently recover the logical relationships among these primitive regions. To achieve accurate primitive region detection, methods like~\cite{Raja2020tabstruct-net,qiao2021lgpma,Zheng2021GTE} leverage the characteristics of cell arrangement to introduce additional constraints. However, these constraints are effective only for tables without deformations. 
%To address more complex table scenarios, primitive regions are described as quadrilaterals, and their corners are detected in~\cite{long2021wtw,xing2023lore}.
For a more precise recovery of the logical positions of primitive regions, FLAG~\cite{liu2021flag} and NCGM~\cite{liu2022NCGM} enhance the feature representation of nodes in the relational graph. LORE~\cite{xing2023lore} employs a cascaded structure to regress the logical position of cells. Limited by cell detection, the performance of these methods in handling blank cells and wireless tables still needs improvement.

\begin{figure*}[htbp]
    \centering
    \includegraphics[width=1\linewidth]{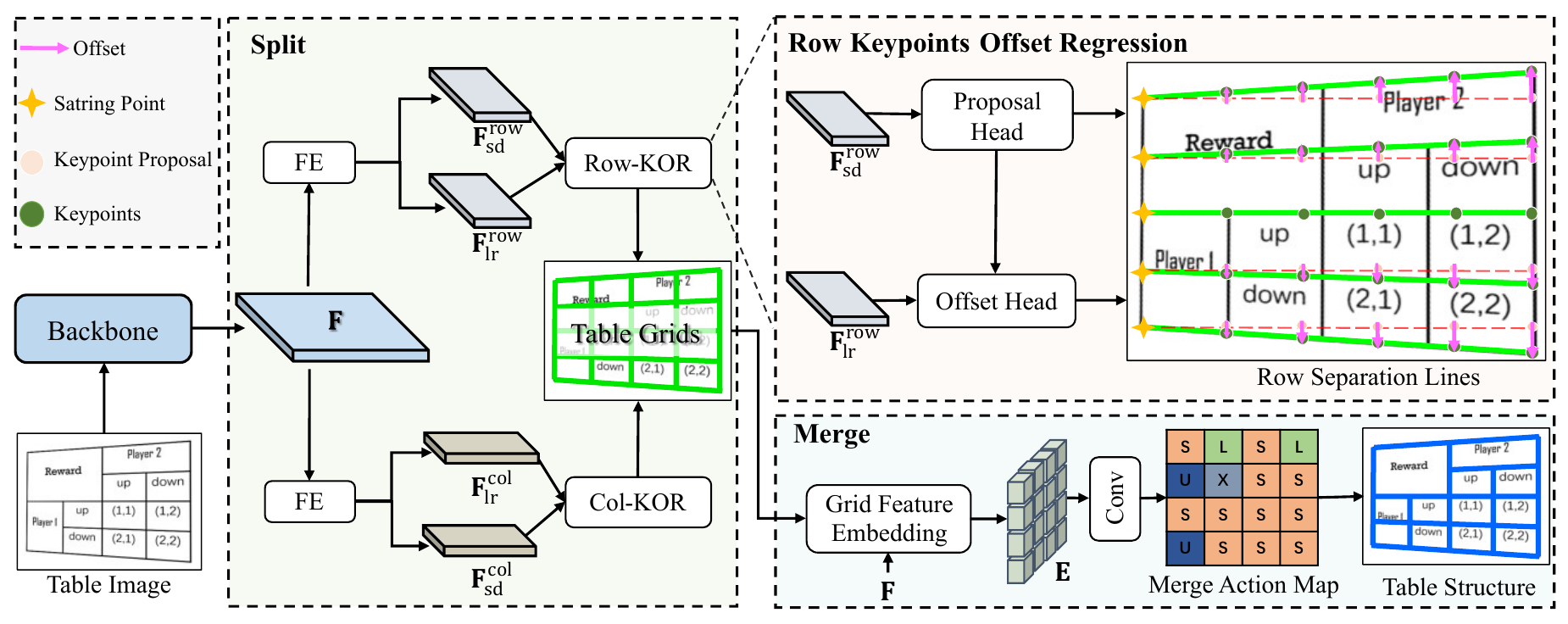}
    \caption{An overview of our approach, SEMv3, which follows the split-and-merge paradigm. In the split stage, we propose the keypoints offset regression module (KOR) to locate the table separation line. The KOR module detects the table separation lines by regressing the offset of each line relative to its keypoint proposals. The ``FE" stands for feature enhancement module. In the merge stage, we utilize a merge action to describe the table structure based on table grids.}
    \label{fig:overview}
\end{figure*}

\subsection{Split-and-merge Based TSR}
Split-and-merge based TSR methods detect table separation lines to parse table grids, then merge the over-split grids into table cells. In the split stage, early methods~\cite{Schreiber2017DeepDeSRT,Tensmeyer2019deep,zhang2022sem} assume that the table lines are aligned, so they could not handle tables with deformations.
In order to address deformed table samples, RobusTabNet~\cite{ma2023RobusTabNet} utilized spatial CNN (SCNN)~\cite{pan2018SCNN} to enhance the feature representation of separation lines. TRACE~\cite{beak2023TRACE} predict the mask of visible and non-visible lines separately. But RobusTabNet and TRACE require complex post-processing of mask-to-line.
SEMv2~\cite{zhang2023semv2} addresses separation line detection in the way of instance segmentation. SEMv2 aggregates instance convolution kernels with global information and then obtains separation line instance-level masks by convolving instance convolution kernels with instance-independent feature maps. However, the position-insensitivity of the instance convolution kernels results in less robust line detection.
TSRFormer~\cite{lin2022tsrformer} is the first to define table separation line detection as line regression and uses Dert to directly regress the position of keypoints. TSRFormer with DQ-DETR~\cite{wang2023tsrformer2} introduces dynamic object queries to gradually regress keypoints. Nevertheless, this approach reduces the efficiency of line regression.
GridFormer~\cite{lyu2023gridformer} completes splitting by directly predicting the coordinates of grid corners.

In the merge stage, RobusTabNet and TSRFormer predict the merging relationship between grids using a relational network~\cite{zhang2017relation}, which do not take advantage of global features of grid cells structure. SEMv2 predicts a merging map for each grid, which is redundant. RNN networks~\cite{cho2014GRU} are used in SEM~\cite{zhang2022sem} to predict cell merging map serially. It is time-consuming and needs additional text information from Off-the-shelf OCR engine.

\section{Method}
% overview
% 主流程图，应该是主图，加线的定义。merge action map 只应该画一张总的。
% 公式的斜体和正体的区别。以及加粗表示向量或者矩阵。
% resnet 和 fpn 不应该
As illustrated in Fig.~\ref{fig:overview}, SEMv3 extracts visual features from the backbone. The split module predicts the row and column separation lines by regressing the offsets between proposals and the keypoints. And then the separation lines are intersected to obtain the table grids. The merge module merges the over-split grids into cells by predicting the merge actions based on starting grid of grids. 

To obtain the visual features of the input table image, we use Resnet-34~\cite{he2016resnet} with FPN~\cite{lin2017FPN} as backbone to generate four different levels of features ${ {\textbf{P}}_2, {\textbf{P}}_3, {\textbf{P}}_4, {\textbf{P}}_5 }$. Subsequently, we amalgamate the information from all levels of the FPN pyramid into a single output $\textbf{F} \in \mathbb{R}^{\frac{H}{2} \times \frac{W}{2} \times C}$. $H$ and $W$ represent the width and height of the input table image, respectively. $C$ is the number of feature channel.

% 分割
\subsection{KOR Based Split Module}
Within the split module, two parallel branches are attached to the shared feature map ${\textbf{F}}$ for the detection of row and column separation lines, respectively. Each branch includes two modules: a feature enhancement module to enrich the context information of ${\textbf{F}}$, and a keypoints offset regression module (KOR) aimed at detecting the separation lines. In the following sections, we will elucidate the details of these two modules by using the row separation line detection branch as an illustrative example.
\label{subsection:split}
% split 和主图融合在一起。
% separation line define
\subsubsection{Feature Enhancement Module}
Inspired by RobusTabNet~\cite{ma2023RobusTabNet}, We employ the Spatial CNN module~\cite{pan2018SCNN} to enhance the feature ${\textbf{F}}$. The spatial CNN module enhances visual features through efficient row and column messaging. We combine repeat 4 down-sample blocks~\cite{lin2022tsrformer} and two cascaded spatial CNN into feature enhancement module. We obtain the row start point detection feature $\textbf{{F}}^{\text{row}}_{\text{sd}} \in \mathbb{R}^{\frac{H}{2} \times \frac{W}{32} \times C}$ and the row line regression feature $\textbf{{F}}^{\text{row}}_{\text{lr}} \in \mathbb{R}^{\frac{H}{2} \times \frac{W}{32} \times C}$ through two row feature enhancement modules with non-shared parameters based on row information passing. Likewise, we calculate enhanced features $\textbf{F}^{\text{col}}_{\text{sd}} \in \mathbb{R}^{\frac{H}{32} \times \frac{W}{2} \times C}$ and $\textbf{F}^{\text{col}}_{\text{lr}} \in \mathbb{R}^{\frac{H}{32} \times \frac{W}{2} \times C}$ for columns.

\subsubsection{Keypoints Offset Regression Module}
% proposals，直接按照行线说明。
\textbf{Separation line representation.} To unify the definition of separation lines in wired tables and wireless tables, we consider the visible cell border line as the separation line in wired tables. For wireless table, we define median line of separation region~\cite{Schreiber2017DeepDeSRT} as row/column separation line. 
We represent the $i^{th}$ row separation line with keypoints $\{{k^{\text{row}}_{ij} | j = 0, ... ,N^{\text{row}}_k-1 }\}$. 
 $N^{\text{row}}_k$ denotes the number of keypoints. $N^{\text{row}}_k$ is determined by the width $W$ of input image and the sampling step size $t$, as defined in Eq.\ref{eq:nk}. 
\begin{align}
    \label{eq:nk}
        N^{\text{row}}_k =  \lceil \frac{W}{t} \rceil
\end{align}

The x-coordinates of the keypoints $k^{\text{row}}_{ij}$ are denoted as $x_{ij}$, where $x_{ij} = j \times t$. Given that the x-coordinates are fixed, the position of a row separation line can be defined solely by the y-coordinates $y$ of its keypoints. To predict these y-coordinates, we regress the y-axis offset $\delta^{\text{row}}_{ij}$ between the keypoints and their proposals. The first keypoint $k^{\text{row}}_{i0}$ in the $i^{th}$ row separation line is designated as the starting point. The keypoint proposals share the same x-coordinates as the keypoints and have y-coordinates identical to the starting point, as shown in the Row Keypoints Offset Regression subplot of Fig.~\ref{fig:overview}.

\textbf{Keypoint proposal head.} The x-axis coordinate of the starting point is fixed at 0. Drawing inspiration from SEMv2~\cite{zhang2023semv2} and TSRFormer~\cite{lin2022tsrformer}, we classify whether each row has a starting point to determine the y-axis coordinate of the starting point of the row separation line.
We predict the probability $\textbf{P}^{\text{row}}$ of each row having a starting point using row-wise average pooling and a softmax operation on the row start point detection feature $\textbf{F}^{\text{row}}_{\text{sd}}$. Subsequently, we apply Non-Maximum Suppression (NMS)~\cite{zhang2023semv2} to eliminate duplicate starting point predictions. 
We then sample $N_k$ keypoint proposals for each row separation line, which are horizontally aligned with the starting point.
%Specifically, we determine the foreground region through a threshold $\theta$. Within the continuous foreground regions, we select the row with the highest probability value.

\label{subsubsection:keypoints refine}
\textbf{Keypoint offset head.} The deformation of table can result in that
keypoint proposals do not accurately represent the curved separation line. To address this challenge, we further predict the offset between the proposals and the ground truth keypoints. The offset helps in refining the location of the keypoints.
We sample features of proposals $\textbf{{K}}^{\prime} \in \mathbb{R}^{C \times N^{\text{row}} \times N^{\text{row}}_k}$ from the row line regression feature $\textbf{{F}}^{\text{row}}_{\text{lr}}$ based on the coordinates of the proposals. Here, $N^{\text{row}}$ indicates the number of row separation lines. We obtain the whole separation line representation features $\textbf{S}^{\text{row}} \in \mathbb{R}^{C \times N^{\text{row}}}$ for each row separation line by averaging the proposals features belonging to the same row, as following: 
\begin{align}
    \label{eq:mean feature}
        \textbf{{S}}^{\text{row}}_i = \frac{1}{N^{\text{row}}_k} \sum_{j=1}^{N^{\text{row}}_k} \textbf{{K}}^{\prime}_{ij}
\end{align}

 To ensure that each keypoint can consider the information of whole line, we concatenate the line representation features of each row to the feature of the proposals in the channel dimension. In Eq.\ref{eq:gird enhance feature}, $\textbf{{K}}$ denotes the keypoint feature.
\begin{align}
    \label{eq:gird enhance feature}
        \textbf{{K}}_{ij} = \text{concatenate}(\textbf{{K}}^{\prime}_{ij}, \textbf{{S}}^{\text{row}}_i)
\end{align}

Instead of using a fully connected network to predict the offsets of keypoints, we employ a $1 \times 3$ convolution layer on the keypoint feature $\textbf{K}$ to predict the offset $\delta_{ij}$ in the row separation line detection branch. This operation allows each keypoint to pay attention to the information of adjacent keypoints, thereby ensuring a smooth separation line.

After predicting row and column separation lines, we create the intersection of the row and column separation lines to obtain quadrilateral boxes of grids $\textbf{{B}} \in \mathbb{R}^{M \times N \times 8}$. $M$ and $N$ denote the number of rows and columns of the grid, respectively. 

\subsection{Merge Module}

% 合并，embedding 和合并一起写。因为没有新的内容。
\subsubsection{Grids Feature Embedding}
The representation feature $\textbf{E}^{\prime} \in \mathbb{R}^{ M \times N \times C_g}$ of the grids is the sum of visual features extracted by the RoiAlign ~\cite{he2017RoIAlign} algorithm and the absolute position embedding (PE)~\cite{xu2020layoutlm} of the grids. $C_g$ is the number of the representation feature channel.
\begin{align}
    \label{eq:extract grids feature}
        \textbf{{E}}^{\prime} = \text{RoiAglin}(\textbf{{F}},\textbf{{B}}) + \text{PE}(\textbf{{B}})
\end{align}
The PE employs a fully connected network to map normalized top-left corner and bottom-right corner coordinates of the grid to $C_g$ dimensions. To enhance $\textbf{E}^{\prime}$, we introduce a row/column-based self-attention mechanism, which automatically aggregates features from other grids, resulting in the enhanced grid representation feature $\textbf{{E}}$.
\begin{figure}
    \centering
    \includegraphics[width=0.9\linewidth]{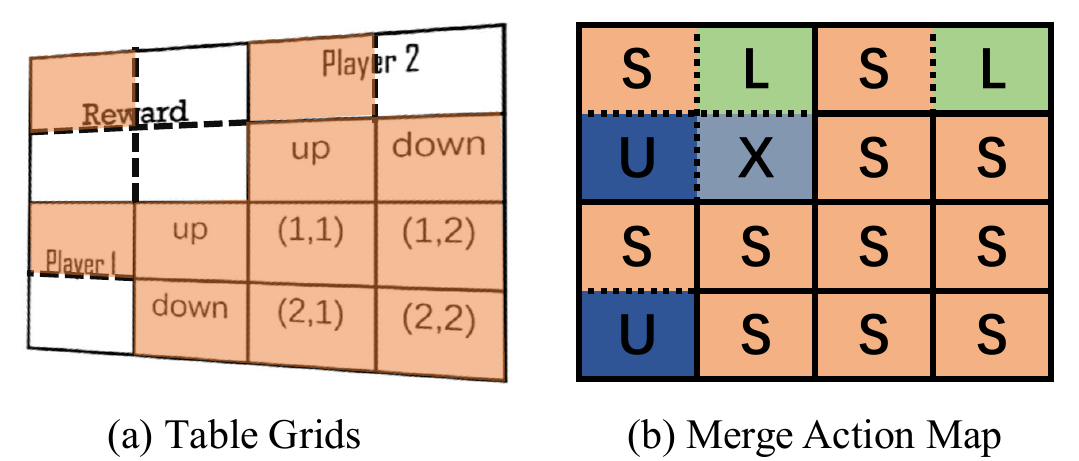}
    \caption{Definition of merge action based on grids. (a) The girds based table structure. The starting grids are painted orange.
The black dashed lines represent the borders within the grid that need to be merged. (b) The merge action map.}
    \label{fig:Merge action definition.}
\end{figure}
\label{subsection:merge}
\subsubsection{Merge Action Prediction}
Inspired by OTSL~\cite{lysak2023optimized} and Formerge~\cite{nguyen2023formerge}, we define the merging action of a grid to depict the table structure, as shown in Fig.~\ref{fig:Merge action definition.}. The upper-left grid in a cell serves as the starting grid, as illustrated in Fig~\ref{fig:Merge action definition.}(a). As shown in Fig.~\ref{fig:Merge action definition.}(b), the merging action for starting grids is \textbf{S}tay, implying that these grids do not need to merge either to the left or upwards. The merge action for grids in the same row as the starting grids is \textbf{L}eft, suggesting that these grids need to merge with the ones on their left. The merge action for grids in the same column as the starting grids is \textbf{U}pward, indicating that these grids need to merge with the ones above them. The merge action for the remaining grids in the same cell is \textbf{X}, which denotes merging both upwards and to the left.
This novel definition of merge action allows us to succinctly and comprehensively characterize the grid based table structure with only one merge action map, as shown in Fig.~\ref{fig:Merge action definition.}(b).

According to the definition of merge action, the starting grids play a pivotal role in determining the merge action of the remaining grids within the same cell. 
To enhance the accuracy of merge action prediction, we introduce an auxiliary branch for the classification of starting grids. This helps in identifying the starting grids during the prediction of merge action. In this branch, we transform $\textbf{E}$ into the starting grid classification feature, $\textbf{E}_s$, using a convolutional layer. $\textbf{E}_s$ is used to classify the starting grids. Finally, we concatenate $\textbf{E}_s$ and $\textbf{E}$ channel-wise to predict the merge action.

\subsection{Training Object}
\label{subsection:training}
\textbf{Starting points detection loss.}
We supervise the prediction of starting points using binary cross-entropy loss ${L}_{\text{BCE}}$. This is represented as:
\begin{align}
    \label{eq:starting points loss}
        \mathcal{L}_{\text{sp}}^{\text{row}} = \frac{1}{H/2} \sum_{i=1}^{H/2} {{L}_{\text{BCE}}(\hat{\textbf{{P}}_i}^{\text{row}},\textbf{{P}}^{\text{row}}_i)}
\end{align}

Although the starting point has a width of only 1 pixel, we enlarge the range of starting points for ease of training. Here, $\hat{\textbf{{P}}}^{\text{row}}_i$ represents the ground truth distribution of starting points for the $i^{th}$ row separation line. In wireless tables, $\hat{\textbf{{P}}}^{\text{row}}_i$ is set to 1 if the $i^{th}$ row is located in the separation region. For wired tables, $\hat{\textbf{{P}}}^{\text{row}}_i$ is 1 if the $i^{th}$ row is located in the neighborhood of starting points with a width of 8.
We also calculate the loss $\mathcal{L}_{\text{sp}}^{\text{col}}$ for column starting points.

\textbf{Keypoints offset regression loss.}
$\hat{\delta}^{\text{row}}_{ij} $ denotes to the ground truth offset between the $j^{th}$ keypoint proposal in the $i^{th}$ row separation line. The loss of row line keypoints offset is calculated with L2 loss, as:
\begin{align}
    \label{eq:delta loss}
        \mathcal{L}_{\delta}^{\text{row}} =  \frac{1}{N^{\text{row}}  N_k^{\text{row}} }  \sum_{i=1}^{N^{\text{row}}} \sum_{j=0}^{N_k^{\text{row}}-1} \Vert \hat{\delta}^{\text{row}}_{ij} - {\delta}^{\text{row}}_{ij} \Vert_2
\end{align}

We define the column line keypoints offset loss $\mathcal{L}_{\delta}^{\text{col}}$ in the same manner.

\textbf{Merge action classification loss.}
We use focal loss in in both predicting starting grids and merge actions.
\begin{align}
    \label{eq:start grid loss}
    \mathcal{L}_{\text{sg}} = \frac{1}{N  M} \sum_{i=1}^{N} \sum_{j=1}^{M} {L}_{\text{focal}}(\hat{\textbf{{P}}}^{\text{sg}}_{ij},{\textbf{{P}}}^{\text{sg}}_{ij}).
\end{align}
\begin{align}
    \label{eq:merge action loss}
    \mathcal{L}_{\text{ma}} = \frac{1}{N  M} \sum_a \sum_{i=1}^{N} \sum_{j=1}^{M} {L}_{\text{focal}}(\hat{\textbf{{P}}}^{a}_{ij},{\textbf{{P}}}^{a}_{ij}).
\end{align}

The loss for starting grid classification and merge action prediction are denoted as $\mathcal{L}_{\text{sg}}$ and $\mathcal{L}_{\text{ma}}$ respectively. $\hat{\textbf{{P}}}^{\text{sg}}$ is set to 1 if a grid is a starting grid. The symbol $a$ represents a merge action, which can take one of four values: \textbf{S}, \textbf{L}, \textbf{U}, or \textbf{{X}}. The term ${L}_{\text{focal}}$ refers to the focal loss~\cite{lin2017focal}.

\textbf{Overall loss.} All the modules in SEMv3 are trained end-to-end. The overall loss function is a summation of several component losses, as shown below:
\begin{align}
    \mathcal{L} = \mathcal{L}_{\text{sp}}^{\text{row}} +
    \mathcal{L}_{\text{sp}}^{\text{col}} + 
    \mathcal{L}_{\delta}^{\text{row}} + 
    \mathcal{L}_{\delta}^{\text{col}} + 
    \mathcal{L}_{\text{ma}} + 
    \mathcal{L}_{\text{sg}}
\end{align}

\section{Experiment}
\subsection{Implementation}
We use Resnet-34 pretrained on ImageNet~\cite{2012ImageNet} with FPN as backbone. And the feature $F$ channel number $C$ is 256, grid feature channel number $C_g$ is 512. The sampling step size $t$ is 32. Models are trained end-to-end for 100 epochs. We use Adam~\cite{2014Adam} as the optimizer. The initial learning rate is $1\times 10^{-4}$, and the learning rate is adjusted to $1\times 10^{-6}$ according to the cosine annealing strategy~\cite{loshchilov2016sgdr}. All experiments are implemented in Pytorch v1.7.1 and conducted on 4 Nvidia Tesla V100 GPUs with 24GB RAM memory. During the training process, the ground truth grid boxes coordinates are used when extracting grid represent feature using RoIAlign.

%The threshold $\theta$ in starting points detection is 0.5.
% 可视化图片排版。同一个系列用同样的字母标号。
\begin{figure*}
    \centering
    \includegraphics[width=1\linewidth]{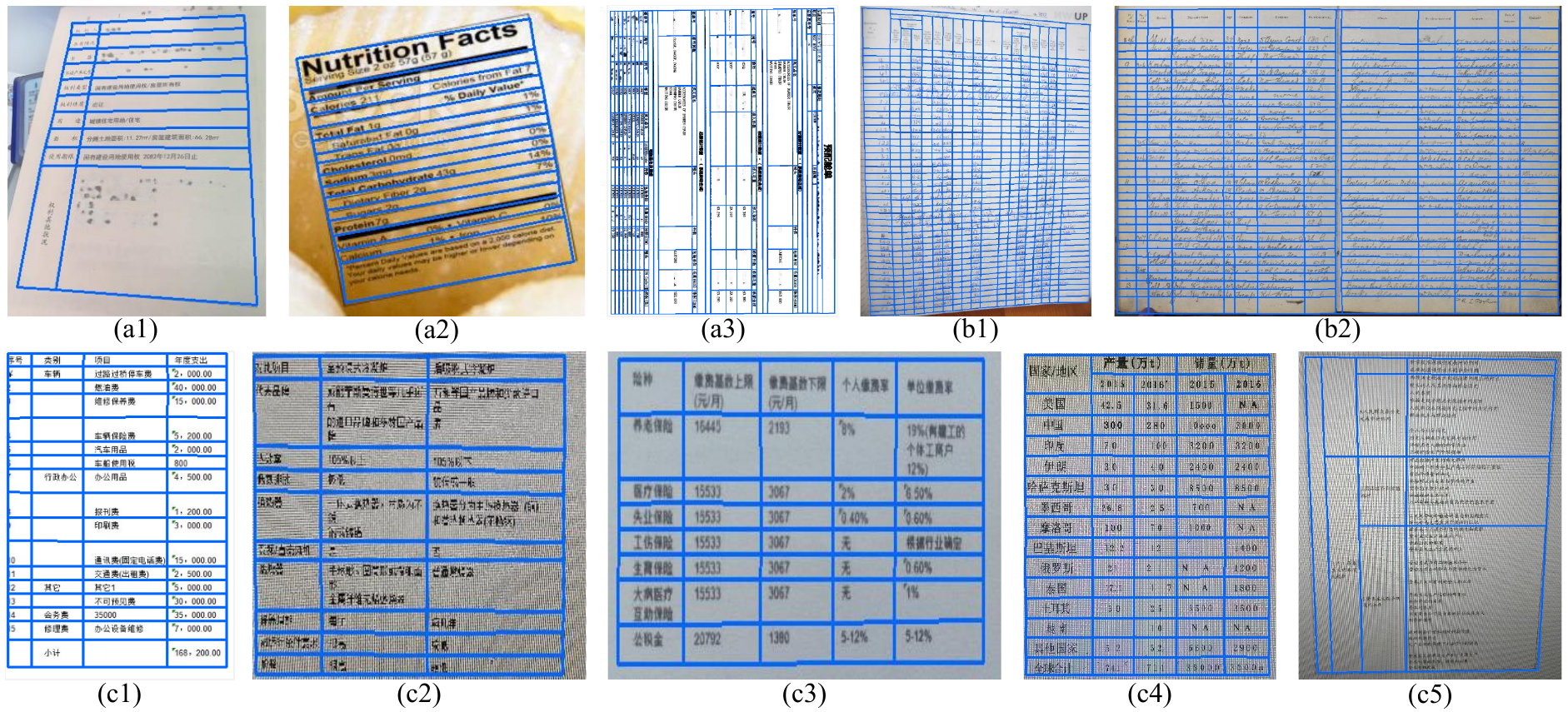}
    \caption{Qualitative table structure recognition results of our approach. (a1-a3) are from WTW. (b1) and (b2) are from ICDAR-2019 cTDaR Historical. (c1-c4) are from the wired subset of iFLYTAB. (c5) is from the wireless subset of iFLYTAB.}
    \label{fig:wtw iFLYTAB result}
\end{figure*}

\begin{table}
    \centering
    \begin{tabular}{lccc}
    \toprule
       Methods&P&R&F1 \\
    \midrule
    TabStructNet~\shortcite{Raja2020tabstruct-net}&82.2&78.7&80.4\\
    FLAGNet~\shortcite{liu2021flag}&85.2&83.8&84.5\\
    NCGM~\shortcite{liu2022NCGM}&84.6&86.1&85.3\\
    SEMv2~\shortcite{zhang2023semv2}&89.2&82.8&85.9\\
    LORE~\shortcite{xing2023lore}&87.9&\textbf{88.7}&88.3\\
    SEMv3&\textbf{90.3}&88.4&\textbf{89.3}\\
    \bottomrule
    \end{tabular}
    \caption{Comparsion with state-of-the-art methods on ICDAR-2019 cTDaR Historical.}
    \label{tab:icdar-2019 result}
\end{table}

\begin{table}
    \centering
    \begin{tabular}{lccc}
    \toprule
       Methods&P&R&F1 \\
    \midrule
    Cycle-CenterNet~\shortcite{long2021wtw}&93.3&91.5&92.4\\
    NCGM~\shortcite{liu2022NCGM}&93.7&94.6&94.1\\
    TSRFormer-DQ~\shortcite{wang2023tsrformer2}&94.5&94.0&94.3\\
    SEMv2~\shortcite{zhang2023semv2}&93.8&93.4&93.6\\
    $\text{LORE}^{*}$~\shortcite{xing2023lore}&94.5&\textbf{95.9}&95.1\\
    SEMv3&\textbf{94.8}&95.4&\textbf{95.1}\\
    \bottomrule
    \end{tabular}
    \caption{Comparsion with state-of-the-art methods on WTW. The method with * means the IoU of cell matching is 0.5 when evoluting.}
    \label{tab:wtw result}
\end{table}

\subsection{Datasets and Metric}
% overview of dataset and Metric
We evaluate the performance of our method on several public datasets. These datasets encompass a wide range of challenging scenarios related to table structure recognition.

\textbf{ICDAR-2019 cTDaR Historical}~\cite{Gao2019icdar2019} dataset contains 600 training samples and 150 testing samples from archival historical documents. This dataset presents several challenges, including a limited number of training sets, unclear row and column separation lines, and deformation in the table area. We use the cell adjacency relationship (IoU=0.6)~\cite{G2012icdar2013,Gao2019icdar2019} as the evaluation metric of this dataset.

\textbf{WTW}~\cite{long2021wtw} contains 14581 wired table images collected from real business scenarios. There are seven difficult cases in this dataset: inclined tables, curved tables, occluded or blurred tables, tables with extreme aspect ratios, overlaid tables, multi-colored tables, and irregular tables. Metric on WTW is the cell adjacency relationship (IoU=0.6)~\cite{G2012icdar2013,Gao2019icdar2019}.

\textbf{iFLYTAB}~\cite{zhang2023semv2} contains 12,103 training samples and 5,188 testing samples. The dataset includes a variety of complex examples from both electronic documents and natural scenes. Based on the table data collection scenarios and table characteristics, iFLYTAB is further divided into four subsets: Wired-Digital (WDD), Wired-Camera-Capture (WDC), Wireless-Digital (WLD), and Wireless-Camera-Capture (WLC). 
 We use the cell adjacency relationship~\cite{G2012icdar2013} followed SEMv2~\cite{zhang2023semv2} as the evaluation metric. To evaluate the performance of the split stage, we use grid detection with an IoU of 0.9~\cite{long2021wtw} as the metric. F1-G represents the F1-score for grid detection, in Table~\ref{tab:ablation}.

\textbf{SciTSR} ~\cite{chi2019scitsr} and \textbf{PubTabNet} ~\cite{zhong2020pubtabnet} are from scientific literature PDF, where table samples have aligned axis. The test subset of SciTSR, SciTSR-comp, is composed of 716 complex tables containing span-cells. We evaluate our method with the cell adjacency relationship metric ~\cite{G2012icdar2013} on SciTSR-comp. TEDS and TEDS-Struct ~\cite{zhong2020pubtabnet} is used to evaluate performance on PubTabNet.

\subsection{Comparison with State-of-the-arts}
Our method focuses on robustly recovering the table structure from deformed and wireless tables in challenging scenarios. On the following three challenging datasets, our method achieves SOTA. On ICDAR-2019 cTDaR Historical dataset, we achieve 88.9\% F1 score, as shown in Table \ref{tab:icdar-2019 result}. On WTW dataset, we achieve performance with SOTA, as shown in Table \ref{tab:wtw result}. Notably, LORE who is the second best on these two datasets takes 0.5 for IoU when evaluating models while we take 0.6. On iFLYTAB dataset, as shown in Table \ref{tab:iFLYTAB result}, we improve SEMv2 by 0.9\% F1 score. As shown in Fig.~\ref{fig:wtw iFLYTAB result}, Our approach performs well on challenging table samples, e.g., deformed tables, blurred tables, wireless tables.

\begin{table}
    \centering
    \begin{tabular}{lccc}
    \toprule
       Methods&P&R&F1 \\
    \midrule
    SEM~\shortcite{zhang2022sem}&81.7&74.5&78.0\\
    SEMv2~\shortcite{zhang2023semv2}&93.8&93.3&93.5\\
    SEMv3&\textbf{94.4}&\textbf{94.2}&\textbf{94.4}\\
    \bottomrule
    \end{tabular}
    \caption{Comparsion with state-of-the-art methods on iFLYTAB.}
    \label{tab:iFLYTAB result}
\end{table}
Our model also demonstrates comparable performance to SOTA in simpler scenarios involving electronic documents. However, the testing dataset of SciTSR has been manually rectified in various ways, resulting in an unfair comparison on the SciTSR dataset. On the PubTabNet, the performance of our method is slightly inferior to that of LORE, which belongs to the region-based methods. In the split-and-merge based methods, we achieve performance comparable to SOTA.

\subsection{Ablation Studies}
\begin{table}
\begin{tabular}{@{}lcccccccc@{}}
\toprule
\multirow{2}{*}{Methods} & \multicolumn{3}{c}{SciTSR-COMP}& \multicolumn{2}{c}{PubTabNet} \\
\cmidrule(lr){2-4} \cmidrule(lr){5-6}
&P&R&F1&TEDS&TEDS-S\\
\midrule RobusTabNet&99.0&98.4&98.7&-&97.0\\
 TSRFormer&99.1&98.7&98.9&-&97.5\\
 LORE&\textbf{99.4}&\textbf{{99.2}}&\textbf{99.3}&\textbf{98.1}&-\\
 TSRFormer-DQ&99.1&98.6&98.8&-&\textbf{97.5}\\
 SEMv2&98.7&98.6&98.7&-&97.5\\
 SEMv3&99.1&98.9&99.0&97.3&\textbf{97.5}\\
\bottomrule
\end{tabular}
\centering
\caption{Comparsion with state-of-the-art methods on SciTSR and PubTabNet in the digital document scenario. TEDS-S stands for TEDS-Struct.}
\label{tab:digital document tables results}
\end{table}

\begin{table*}[t]
\begin{tabular}{@{}cccccccccccccccc@{}}
\toprule
\multirow{3}{*}{System}& \multicolumn{2}{c}{Split} &\multicolumn{2}{c}{Merge} & \multicolumn{2}{c}{WLC}          & \multicolumn{2}{c}{WLD}         & \multicolumn{2}{c}{WDC} & \multicolumn{2}{c}{WDD} & Total \\ \cmidrule(lr){2-3} \cmidrule(lr){4-5} \cmidrule(lr){6-7} \cmidrule(lr){8-9} \cmidrule(lr){10-11} \cmidrule(lr){12-13} \cmidrule(l){14-14}  
&   IS & KOR &  MP & MA& F1 & F1-G                      & F1                      & F1-G & F1       & F1-G       & F1       & F1-G       & F1  \\ \midrule
T1&\checked & $\times$ &  \checked & $\times$& 90.8 & 18.5                         & 92.4                      & 24.0    & 95.3       & 59.1          & 95.5       & 51.8          & 93.3  \\
T2&$\times$ & \checked &  \checked & $\times$& 91.7 & 37.1                         & 92.8                      & 53.7    & 95.5       & \textbf{64.6}          & 95.3       & \textbf{63.6}          & 93.6  \\ 
T3&\checked & $\times$ &  $\times$ & \checked& 91.6 & 20.4                         & 93.1                      & 27.2    & \textbf{96.5}       & 60.6          & \textbf{96.3}       & 53.0          & 94.0  \\
T4&$\times$ & \checked &  $\times$ & \checked& \textbf{92.5} & \textbf{37.9}                         & \textbf{93.4}                      & \textbf{59.9}    & 96.4       & \textbf{64.6}          & \textbf{96.3}       & 63.2          & \textbf{94.4}  \\ \bottomrule
\end{tabular}
\centering
\caption{The ablation experiments of the split module and merge module on the iFLYTEK and its different subsets. ``KOR" refers to the keypoints offset regression-based method proposed in this paper, while ``IS" represents the instance segmentation-based split module from SEMv2. ``MA" is the new merge module introduced in this paper, and ``MP" is the merge module from SEMv2. The F1 score used here is a metric for cell adjacency relationships. The F1-G score evaluates grid detection and reflects the performance of table separation line detection in the split stage.}
\label{tab:ablation}
\end{table*}

To verify the effectiveness of our split and merge module, we undertake ablation experiments through several systems designed as shown in Table \ref{tab:ablation}. In the split stage, the Instance Segmentation (IS) based separation line detection module from SEMv2, employs dynamic convolution to generate a segmentation mask for each separation line. We add a new convolutional layer to generate instance-independent features, following the original SEMv2 settings. In this paper, we introduce the KOR module to detect separation lines by regressing the offset between the key points and their proposals. In the merge stage, we propose a merge action based merging module(MA). Also, we implemented the ParaDec merge module in SEMv2 to predict the merge map (MP) for each grid.

\subsubsection{The Robustness of the Split Module}
\begin{figure}
    \centering
    \includegraphics[width=1\linewidth]{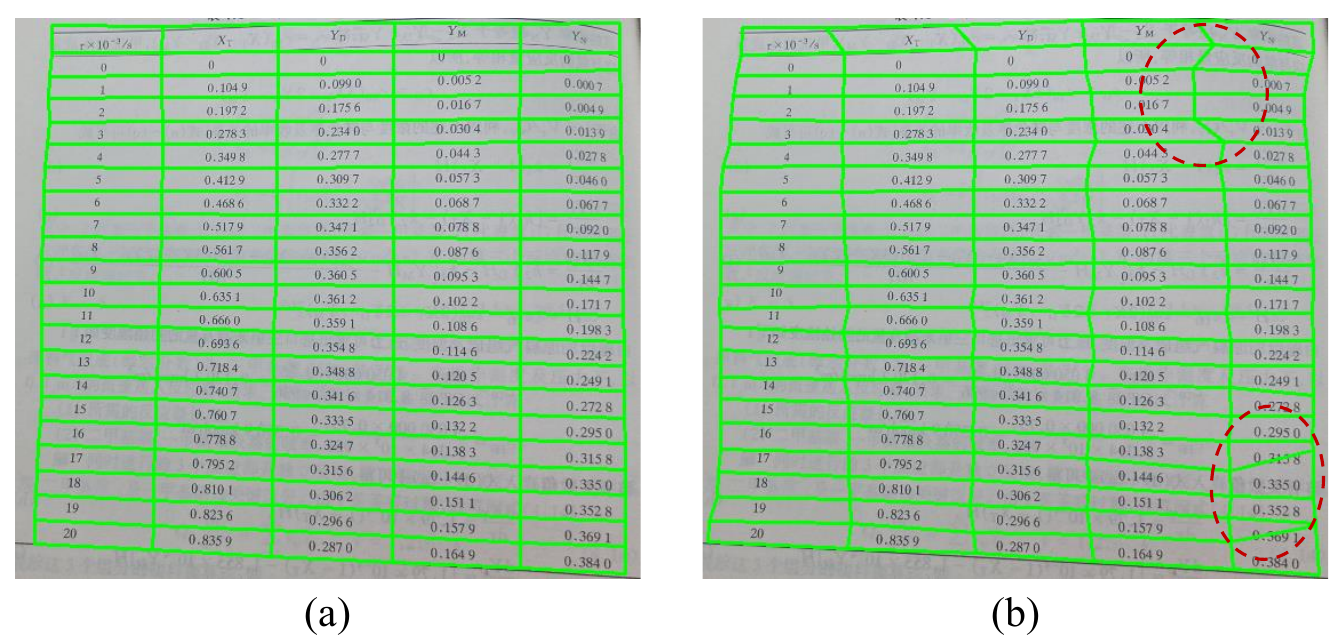}
    \caption{Qualitative split results of different split modules. (a) is the prediction of table grids from KOR, (b) is the prediction of table grids of table grids from IS. The red dashed box indicates low-quality results.}
    \label{fig:split result}
\end{figure}

To establish the robustness of our split module, we compare the performances of IS and KOR under same merge modules. T1 and T2 serve to compare the performance of IS and KOR under the MP merge module. On the other hand, T3 and T4 illustrate the performance of IS and KOR under the MA merge module. Under both merge module settings, our KOR consistently achieves superior split results, as evidenced by the F1-G. This is particularly notable with the wireless tables (WLD and WLC), where KOR performs almost twice as well as IS. As depicted in Fig.~\ref{fig:split result}, the split results of T1 and T2 on the wireless tables clearly demonstrate that KOR produces split lines of higher quality. Furthermore, our KOR proves to be more stable compared to the IS split results when the merge module changes.

\subsubsection{The Speed of the Split Module}
\begin{figure}
    \centering
    \includegraphics[width=0.9\linewidth]{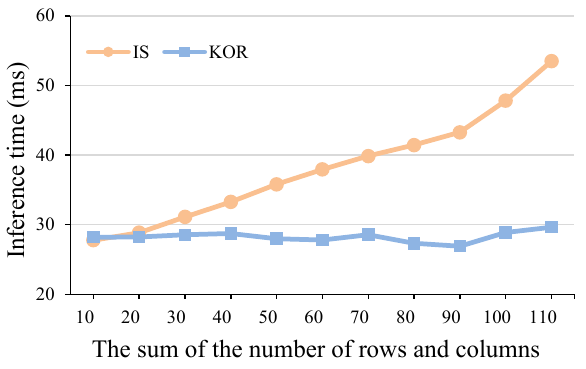}
    \caption{Comparison of inference time for different split modules.}
    \label{fig:speed}
\end{figure}

The IS split module generates a mask for each table separation line, leading to a computational cost that increases linearly with the number of rows and columns. Conversely, our method consistently represents the position of lines with a single offset map, meaning the computational cost does not significantly increase with the growth in rows and columns. We evaluated the inference time of both split modules on the SciTSR test dataset, with images resized to $512 \times 512$. As illustrated in Fig.~\ref{fig:speed}, the x-axis represents the sum of the number of rows and columns with a step of 10, while the y-axis represents the inference time. The inference time of our KOR is less affected by the increase in the number of rows and columns. Particularly in cases with a large number of rows and columns, KOR is significantly faster than IS.

\subsubsection{The Effectiveness of the Merge Module}

As shown in Table \ref{tab:ablation}, MA always outperforms MP when the split module is kept constant. MA have a positive gain on the detection accuracy of separation lines with IS.
Our merge module can significantly reduce the memory occupancy. The space complexity of MA is $\mathcal{O}(NM)$, but MP has a space complexity of $\mathcal{O}(N^2M^2)$. Benefiting from our modeling of the merge action, we can merge over-split grids with high accuracy using a much simpler network structure than SEMv2. 

\section{Conclusion}
In this paper, we introduce SEMv3, a fast and robust table structure recognizer that adheres to the split-and-merge paradigm. In the split stage, we propose a split module (KOR) based on keypoints offset regression for robust and fast detection of split lines. KOR detects separation lines by regressing the offset between the keypoint proposals and the line. During the merge phase, we define a grid-based merge action to effectively characterize the table structure. Our method achieves state-of-the-art performance on the ICDAR-2019 cTDaR Historical, WTW, iFLYTAB datasets, demonstrating its effectiveness on table samples in challenging scenarios, such as wireless tables and deformed tables. Ablation experiments show that our split module KOR is fast, robust, and achieves significant improvement on wireless tables. Our merge module is also efficient.
\section*{Contribution Statement}
Chunxia Qin: Conceptualization, Methodology, Writing - original draft,review and editing. Zhenrong Zhang: Conceptualization, Methodology, Writing -review and editing. Pengfei Hu: Writing -review and editing. Chenyu Liu: Methodology, Writing -review and editing, Resources. Jiefeng Ma: Writing -review and editing. Jun Du: Methodology, Writing - review and editing.

\bibliographystyle{named}
\bibliography{ijcai24}

\end{document}